\documentclass{article}

    \PassOptionsToPackage{numbers, compress}{natbib}

\usepackage[dvipsnames, table]{xcolor} 

\usepackage[preprint]{neurips_2025}

\usepackage[utf8]{inputenc} 
\usepackage[T1]{fontenc}    
\usepackage{hyperref}       
\usepackage{url}            
\usepackage{booktabs}       
\usepackage{amsfonts}       
\usepackage{nicefrac}       
\usepackage{microtype}      
\usepackage{xcolor}         

\usepackage{algorithm}
\usepackage{algpseudocode}
\usepackage{subcaption} 
\usepackage{wrapfig,lipsum,booktabs}
\usepackage{multirow}
\usepackage{blindtext}
\usepackage{tcolorbox}
\usepackage{graphicx} 
\usepackage{pifont}
\usepackage{amsmath}
\usepackage{enumitem}
\usepackage{hyperref}
\hypersetup{
colorlinks = true,
breaklinks=true,
colorlinks=true,
citecolor=blueish,
linkcolor=red,
}

\usepackage[capitalize]{cleveref}
\crefname{section}{Sec.}{Secs.}
\Crefname{section}{Section}{Sections}
\Crefname{table}{Table}{Tables}
\crefname{table}{Tab.}{Tabs.}

\definecolor{maroon}{cmyk}{0,0.87,0.68,0.32}
\definecolor{myyellow}{RGB}{218, 160, 109}
\definecolor{brickred}{rgb}{0.8, 0.25, 0.33}
\definecolor{brandeisblue}{rgb}{0.0, 0.44, 1.0}
\definecolor{applegreen}{rgb}{0.55, 0.71, 0.0}
\definecolor{aogreen}{rgb}{0.0, 0.5, 0.0}

\definecolor{gdmb}{RGB}{47, 114, 173}  
\definecolor{gdmr}{RGB}{199, 100,  38}
\definecolor{gdmg}{RGB}{70, 155, 118}
\definecolor{gdmm}{RGB}{193, 126, 165}
\definecolor{gdmy}{RGB}{239, 227,  98}
\definecolor{gdmc}{RGB}{110, 179, 228}
\definecolor{gdmk}{RGB}{20, 20, 20}
\definecolor{turquoise}{cmyk}{0.65,0,0.1,0.3}
\definecolor{purple}{rgb}{0.65,0,0.65}
\definecolor{dark_green}{rgb}{0, 0.5, 0}
\definecolor{orange}{rgb}{0.8, 0.6, 0.2}
\definecolor{red}{rgb}{0.8, 0.2, 0.2}
\definecolor{darkred}{rgb}{0.6, 0.1, 0.05}
\definecolor{blueish}{rgb}{0.0, 0.3, .6}
\definecolor{light_gray}{rgb}{0.7, 0.7, .7}
\definecolor{pink}{rgb}{1, 0, 1}
\definecolor{greyblue}{rgb}{0.25, 0.25, 1}
\definecolor{orgred}{rgb}{1.0, 0, 0}

\usepackage{pifont}

\definecolor{sh_gray}{rgb}{0.84,0.84,0.84}
\definecolor{sh_gray2}{rgb}{1,0.89,0.75}
\definecolor{color3}{rgb}{0.95,0.95,0.95}
\definecolor{color4}{rgb}{0.94,0.94,1}
\definecolor{color5}{rgb}{1,0.96,0.88}

\title{\textit{DINO-R1}: Incentivizing Reasoning Capability in Vision Foundation Models}

\author{
  Chenbin Pan$^{1,2}$ \quad Wenbin He$^{1,2}$ \quad Zhengzhong Tu$^{3}$ \quad Liu Ren$^{1,2}$ \\
  $^{1}$Bosch Research North America \\
  $^{2}$Bosch Center for Artificial Intelligence (BCAI) \\
  $^{3}$Texas A\&M University \\
  \texttt{\{chenbin.pan@us.bosch.com\}} \\
  \vspace{0.5em}
  \href{https://Christinepan881.github.io/DINO-R1}{\texttt{https://Christinepan881.github.io/DINO-R1}}
}

\begin{document}

\maketitle

\begin{abstract}
The recent explosive interest in the reasoning capabilities of large language models, such as DeepSeek-R1, has demonstrated remarkable success through reinforcement learning-based fine-tuning frameworks, exemplified by methods like Group Relative Policy Optimization (GRPO). However, such reasoning abilities remain underexplored and notably absent in vision foundation models, including representation models like the DINO series. In this work, we propose \textbf{DINO-R1}, the first such attempt to incentivize visual in-context reasoning capabilities of vision foundation models using reinforcement learning. Specifically, DINO-R1 introduces \textbf{Group Relative Query Optimization (GRQO)}, a novel reinforcement-style training strategy explicitly designed for query-based representation models, which computes query-level rewards based on group-normalized alignment quality. We also apply KL-regularization to stabilize the objectness distribution to reduce the training instability. This joint optimization enables dense and expressive supervision across queries while mitigating overfitting and distributional drift. Building upon Grounding-DINO, we train a series of DINO-R1 family models that integrate a visual prompt encoder and a visual-guided query selection mechanism. Extensive experiments on COCO, LVIS, and ODinW demonstrate that DINO-R1 significantly outperforms supervised fine-tuning baselines, achieving strong generalization in both open-vocabulary and closed-set visual prompting scenarios.
\end{abstract}

\section{Introduction}
\label{sec:introduction}

Recent advancements in large reasoning models (LRMs) \cite{grpo,llama2,llama3,qwen,qwen2,bi2024deepseek,lu2024deepseek,guo2025deepseek,liu2024deepseek,llava}, exemplified by the impressive performance of DeepSeek-R1 \cite{guo2025deepseek,grpo}, have demonstrated remarkable capabilities across complex reasoning tasks like math reasoning and coding. This breakthrough is largely driven by innovative reinforcement (RL) learning strategies such as Group Relative Policy Optimization (GRPO) \cite{grpo}. By iteratively generating synthetic data and optimizing reasoning models through verifiable rewards, DeepSeek-R1 has attained superior reasoning abilities competitive with state-of-the-art proprietary models, such as OpenAI o1, significantly reshaping the language modeling landscape. Despite these impressive advancements, however, the equivalent advancement of reasoning remains notably limited in vision foundation models (VFMs) \cite{sam,sam2,dino,gdino,dinox,dinov,vit}. Current VFMs largely rely on supervised training paradigms focused on predefined visual categories \cite{imagenet,cocods,o365} or self-supervised objectives \cite{dinov2,SimCLR,G-SimCLR}. These conventional supervised methods inherently lack robust reasoning mechanisms, limiting their ability to effectively generalize to novel, ambiguous, or high-variance scenarios frequently encountered in practical, real-world applications.

One emerging and increasingly important scenario in VFMs is \textit{visual prompting} \cite{trex,dinov,trex2,dinox,cp-detr}, a new paradigm wherein users specify detection targets using visual exemplars. This approach offers substantial practical utility in broad applications such as auto-labeling, industrial inspection, and robotic manipulation. \cite{app-visual-autovp,app-visual-moka,app-visual-multimodal,app-visual-pivot,app-visual-prompt-inds,app-visual-promptcharm,app-visual-prompting-inds,app-visual-survey,app-visual-understanding,app-visual-vp3d} However, training visual-prompting models poses new challenges despite their practical relevance due to the high diversity and intra-class variation among visual exemplars. Compared to their language-prompted counterparts \cite{gdino,gdino1.5,mmgdino,cp-detr,ovd-prompt}, training recipes for visual prompting detectors remain largely underdeveloped. We have observed that training with supervised fine-tuning (SFT) alone typically struggles under these conditions, exhibiting unstable convergence, limited generalization to out-of-domain data, as well as poor alignment of query predictions with visual prompts (Sec.~\ref{sec:experiments}). These findings suggest that vanilla SFT is insufficient for effectively training visual prompting detectors, motivating us to explore fundamentally new training strategies capable of effectively reasoning over diverse visual inputs for robust generalization.

Inspired by the recent breakthroughs in RL-based training frameworks for LRMs \cite{bi2024deepseek,lu2024deepseek,guo2025deepseek,liu2024deepseek,grpo,ppo,rlhf,dpo,llama2,llava,llama3,qwen,qwen2}, which efficaciously exploit large-scale noisy training data, we aim to similarly unlock the potential of reasoning capabilities within pure vision models, \textit{e.g.}, VFMs. Yet, na\"ive application of language-based RL methods such as GRPO to vision presents non-trivial challenges. For one thing, GRPO assumes the model behaves as a probabilistic generator, explicitly sampling diverse outputs from learned distributions per input. In contrast, vision models typically produce deterministic structured predictions, making it nontrivial to optimize over a sampled output space. For another, GRPO's KL-regularization that stabilizes training via constraining token-level output distributions in language models can not be easily translated to structured visual predictions due to fundamental differences in language and vision formulation.

\begin{figure}
    \centering\includegraphics[width=1.0\linewidth]{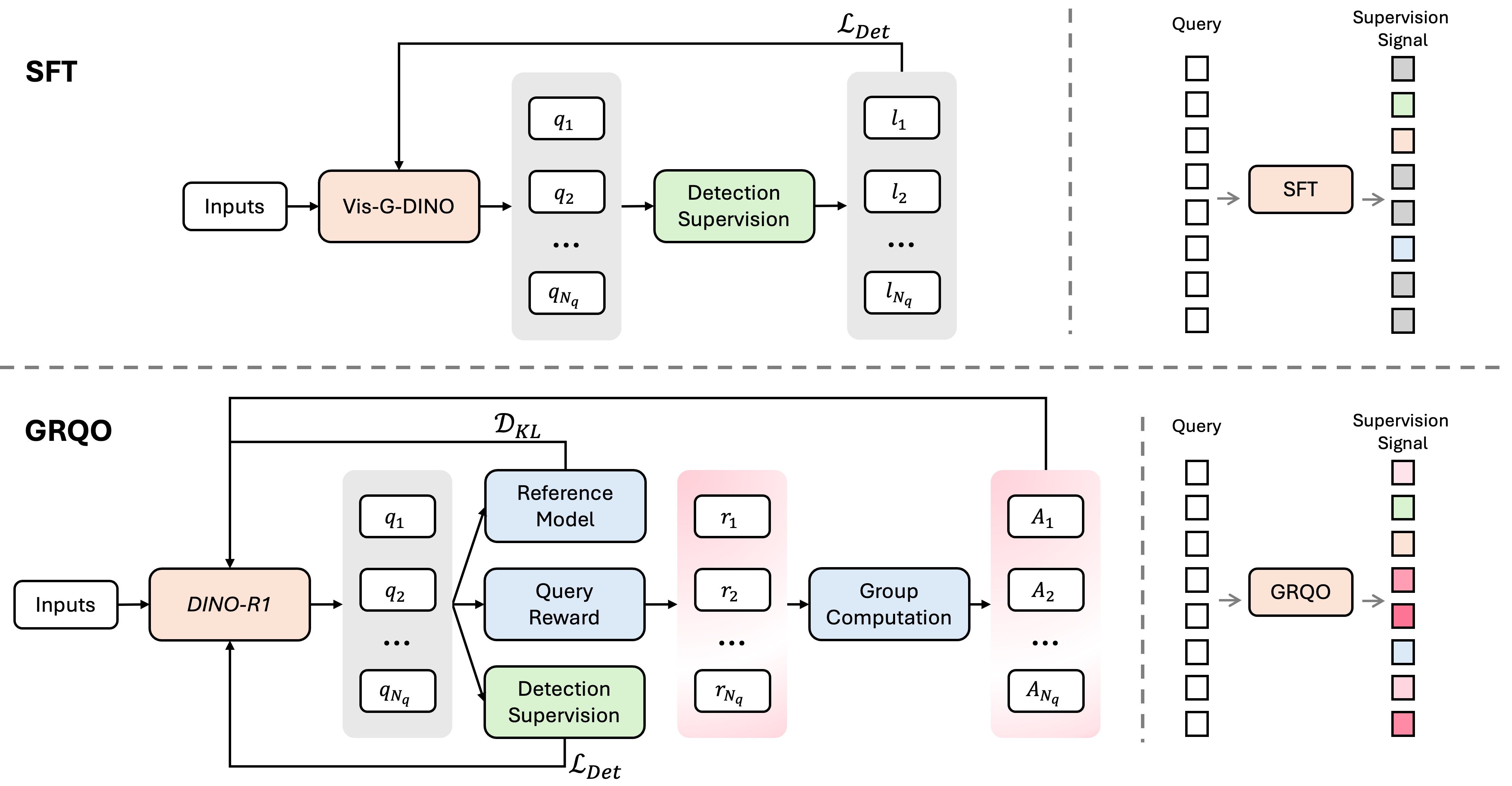}
    \caption{\textbf{SFT vs. GRQO.} SFT leads to limited and homogeneous supervision signals, while GRQO produces richer and more diverse learning signals, encouraging queries to be more expressive.}
    \label{fig:sft-grqo}
\end{figure}

To this end, we present a novel vision-centric RL learning method called group related query optimization (GRQO) designed to incentivize reasoning capabilities in VFMs, notably DINO families. Specifically, GRQO introduces a query-level relative reward module that evaluates each query’s quality within its group and computes a normalized reward based on its advantage over the group average. By encouraging every query to exceed the dynamic group baseline, this mechanism provides denser and more informative training signals (Fig.~\ref{fig:sft-grqo})—in contrast to the traditional one-to-one matching scheme. Additionally, we propose a KL-divergence regularization on the object probability distribution at the query selection stage to help mitigate model drift and catastrophic forgetting during training. To support this new training paradigm, we implemented diverse visual prompting by introducing a visual prompt encoder and a visual-guided query selection mechanism, resulting in a text-free variant we refer to as the \textbf{VIS-G-DINO} baseline. We then train this model using our proposed GRQO framework, which has delivered a set of vision LRMs, here referred to as \textit{DINO-R1} for the final artifact. Our main contributions are summarized as follows:
\begin{itemize}[leftmargin=1.2em,itemsep=1pt]
\vspace{-4pt}
\item We propose \textbf{Group Relative Query Optimization (\textbf{GRQO})}, the first reinforcement-style training paradigm designed to address the high variance of visual prompts in open-set object detection.
\item We introduce a \textbf{query-level relative reward} module and a \textbf{KL-divergence regularization} strategy to improve training stability, query quality, and generalization under visual prompting.
\item We develop \textbf{VIS-G-DINO}, an RL training framework for visual prompt-based VFMs (\textbf{e.g.}, DINO), and define \textbf{\textit{DINO-R1}} as the resulting detector trained with GRQO.
\item We conduct extensive experiments on \textbf{COCO}, \textbf{LVIS}, and \textbf{ODinW}, where DINO-R1 consistently outperforms supervised fine-tuning baselines and demonstrates strong generalization in both open-vocabulary and closed-set visual prompting scenarios.
\end{itemize}

\section{Related Work}
\label{sec:related}

\paragraph{Vision Foundation Models and DETR.}
Vision foundation models (VFMs) \cite{clip,dinov,dinox,florence,gdino,gdino1.5,glip,internvl,sam,sam2,swin,trex,trex2,vlmr1,yoloworld} have enabled significant progress across a wide spectrum of visual tasks by learning general-purpose image representations from large-scale datasets. Among them, DETR \cite{detr} and its derivatives \cite{dndetr,dabdetr,dino,dinov,gdino,ovdetr-cm,dinox,gdino1.5,mmgdino} formulate object detection as a set prediction problem using transformer-based architectures, offering open-set capabilities and strong performance on dense prediction tasks. Grounding DINO \cite{gdino,mmgdino}, in particular, extends DETR to open-vocabulary detection by incorporating vision-language alignment, enabling phrase-grounded object detection via language prompts.

\paragraph{Open-Vocabulary and Prompt-Based Detection.}
Open-vocabulary detection \cite{ovrcnn}(OVD) aims to recognize categories beyond the supervised training set by leveraging external knowledge sources such as pretrained text encoders or image-text pairs. Most existing OVD methods \cite{ovd-distill,ovd-prompt,ovdetr-cm,ovrcnn,ovseg-diff,gdino} focus on language prompts to bridge the category gap, leaving the space of \emph{visual prompting}—using visual exemplars instead of text—largely underexplored. Recent works \cite{cp-detr,dinov,trex,trex2,dinox} have explored using reference images or bounding boxes to ground object-level semantics. However, these methods often rely on inference-time conditioning and lack a robust training paradigm for learning from high-variance visual prompts.

\paragraph{Reinforcement Learning in Foundation Models.}
Reinforcement learning \cite{rlhf,ppo,dpo,grpo,bi2024deepseek} has played a central role in fine-tuning large language models (LLMs)\cite{llama2,llama3,bert,qwen,qwen2,bi2024deepseek,guo2025deepseek} via methods such as Reinforcement Learning from Human Feedback (RLHF) \cite{rlhf} and Group Relative Policy Optimization (GRPO)\cite{grpo,guo2025deepseek}. These approaches enable models to align better with diverse, weakly-supervised, or ambiguous objectives. However, the application of reinforcement-style training to vision foundation models, especially for dense prediction tasks like object detection, remains underexplored. Our work bridges this gap by adapting GRPO principles to query-level learning in transformer-based object detectors through our proposed Group Relative Query Optimization.

\section{Methodology}
\label{sec:method}
While language-prompted object detection has received growing attention in the vision-language community, the training strategy for visual prompt-based detection remains underexplored. To address the challenge of high-variance visual prompts and unlock the potential of prompt-guided detector, in this work, we introduce a novel training paradigm for the visual prompting object detection, Group Relative Query Optimization (GRQO). Built on top of the Grounding-DINO (G-DINO) \cite{gdino} framework (\S \ref{ssec:preliminary}), our approach integrates a visual prompt encoder (\S \ref{ssec:visprompt-enc}) and the GRQO mechanism (\S \ref{ssec:GRQO}) to enhance query learning and improve detection robustness.

\subsection{Preliminary}
\label{ssec:preliminary}

Given an image-text pair $(Image, Text)$, G-DINO extracts multi-scale visual features $\mathbf{I} \in \mathbb{R}^{N_I \times C}$ using an image backbone $\mathcal{B}_{img}$ (e.g., Swin Transformer), and textual features $\mathbf{t} \in \mathbb{R}^{N_{txt} \times C}$ using a text backbone $\mathcal{B}_{txt}$(e.g., BERT).
These features are passed through a cross-modal feature enhancer $\mathcal{F}_{I,t}$ to obtain refined features $\mathbf{I}'$ and $\mathbf{t}'$ via a combination of deformable self-attention (for images), vanilla self-attention (for text), and bidirectional cross-attention for fusion.
To leverage textual guidance during detection, G-DINO employs a language-guided query selection mechanism, where $N_q$ image locations that are most relevant to the text prompt are selected based on cross-modal similarity, and further serving as the positional part for the decoder queries.
A set of learnable queries are conducted to attend to prompts and objects through a multi-modal decoder, which consists of self-attention, image cross-attention, text cross-attention, and feedforward module.
The final class prediction is made through contrastive similarity between queries and refined prompt features.
Following DETR-style supervision, the model is trained using focal loss for classification and a combination of L1 and GIoU loss for bounding box regression. The overall loss is:
\begin{equation}
\mathcal{L}_\text{G-DINO} = \mathcal{L}_\text{focal} + \mathcal{L}_{l1} + \mathcal{L}_\text{GIoU}.
\label{eq:gdino-loss}
\end{equation}

\begin{figure}
    \centering
    \includegraphics[width=1.0\linewidth]{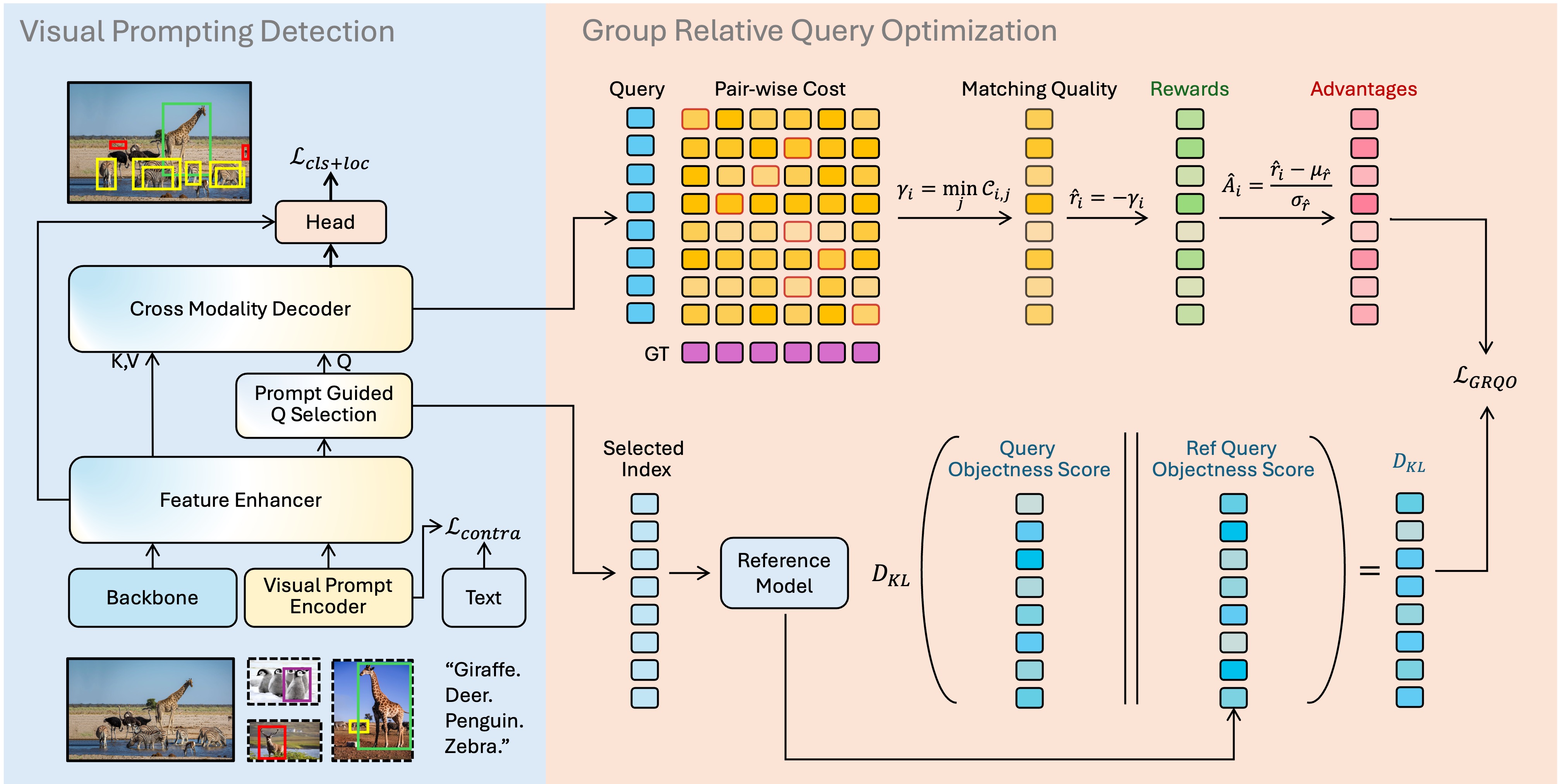}
    \caption{{\bf Overview of the proposed Group Relative Query Optimization (GRQO) framework.}
    The query reward module enriches supervision by assigning group-relative rewards to enhance query learning. In parallel, KL-regularization constrains the trust region of query updates, enabling the detector to progressively absorb diverse visual prompts while preserving previously acquired detection behaviors.
    }
    \label{fig:grpo}
\end{figure}

\subsection{VIS-G-DINO}
\label{ssec:visprompt-enc}
We extend Grounding DINO to support visual prompting and refer to the resulting model as VIS-G-DINO.
Different from G-DINO conditioning on free-form text, VIS-G-DINO conditions detection on \textit{visual prompts}—user-specified bounding boxes over reference images—enabling open-set detection without language description. 
The reference image can be the same as the target image or come from a different context.

\paragraph{Visual prompt encoding.}
We design a visual prompt encoder $\mathcal{E}_{vis}$ that transforms each input bounding box on a reference image into a localized visual feature. Each box is first embedded using sine-cosine positional encoding and projected to match the transformer input space. These embeddings, along with a learnable visual query, are used to attend to multi-scale image features via deformable cross-attention. A self-attention and feedforward layer further refine these into compact visual prompt embeddings that capture region-level semantics. The process can be expressed as: 
\begin{align}
\mathbf{Q}_{\mathcal{E}_{vis}}^{pos} &= \mathrm{Linear}(\mathcal{P}(\mathbf{b_1},...,\mathbf{b_N})) : \mathbb{R}^{N \times 4C} \to \mathbb{R}^{N \times C}, \label{eq:vis-prompt-enc-0} \\
\mathbf{Q}'_{\mathcal{E}_{vis}} &= \mathrm{MSDeformAttn}(\mathbf{Q}_{\mathcal{E}_{vis}}, \mathbf{Q}_{\mathcal{E}_{vis}}^{pos}, \mathbf{b}, \hat{\mathbf{I}} ), \\
\mathbf{v}&=\mathrm{FFN}(\mathrm{SelfAttn(\mathbf{Q}'_{\mathcal{E}_{vis}} )}).
\label{eq:vis-prompt-enc-1}
\end{align}

\paragraph{Semantic Alignment and Prompt Sampling.}
To reinforce semantic consistency, we apply region-level contrastive learning between visual prompts and their corresponding text embeddings. This anchors the visual prompts in the same semantic space as the pretrained language model. During training, we apply random sampling over visual prompts to improve generalization. Specifically, $M$ prompts are randomly sampled per class in a batch to form the final visual instructions $\bar{\mathbf{v}}$. We find $M=1$ yields the best trade-off between diversity and stability.

\paragraph{Image-prompt fusion and query selection.}
Following G-DINO’s architecture, we fuse image features and visual prompts via a multimodal feature enhancer $\mathcal{F}_{I,v}$ to obtain refined image features $\mathbf{I}'$ and visual prompt features $\bar{\mathbf{v}}'$.
To guide detection process, we introduce a \textit{visual-guided query selection} mechanism. Given refined image tokens $\mathbf{I}'$ and visual prompt features $\bar{\mathbf{v}}'$, we compute an image-prompt similarity matrix via dot product. For each image token, we take the maximum similarity across prompt axis as its objectness score, representing the likelihood that a prompted object exists at that location. We select the top-$N_q$ image tokens with the highest objectness scores as positional embeddings for the decoder queries. The top-$N_q$ indices selection can be expressed as:
\begin{equation}
\mathrm{Idx}_{N_q}^{v} = \mathbf{Top}_{N_q}(\mathrm{Max}^{(-1)}(\mathbf{I}' \cdot {\bar{\mathbf{v}}}^{'\top})).
\end{equation}
The corresponding regions serve as coarse proposals, while the content embeddings of the queries remain learnable. The remaining stages mirror the G-DINO pipeline.

\paragraph{Overall training objective.}
The VIS-G-DINO model is optimized using a composite objective:
\begin{equation}
\mathcal{L}_\text{VIS-G-DINO} = \mathcal{L}_\text{contra} + \mathcal{L}_\text{focal} + \mathcal{L}_{L1} + \mathcal{L}_\text{GIoU},
\end{equation}
where $\mathcal{L}_\text{contra}$ promotes semantic alignment, and the remaining losses follow standard detection objectives for classification and localization.

\subsection{Group Relative Query Optimization}
\label{ssec:GRQO}

Visual prompting detection demands that object queries align with highly diverse visual exemplars that share the same semantics. This setting introduces greater intra-class variance than language prompts, requiring the model to both memorize diverse appearances and generalize across unseen variations. Inspired by the generalization ability of GRPO \cite{grpo} in LLMs community, we propose Group Relative Query Optimization (GRQO, Fig.\ref{fig:grpo})—a novel training paradigm that enhances query quality and learning stability through group-based reward modeling and distributional regularization.

\paragraph{Query-level relative reward.}
In DETR-style architectures, queries interact via self- and cross-attention across layers and serve as the main carriers of detection capacity. However, the standard one-to-one bipartite matching provides sparse supervision, updating only a small subset of queries and leaving others under-optimized (Fig.\ref{fig:sft-grqo}). To address this, we introduce a query-level reward mechanism that densifies supervision across all queries. 
Instead of relying solely on bipartite matching to backpropagation gradients to a limited subset of queries, we compute an auxiliary reward signal for each query based on its alignment quality with ground-truth instances.
Specifically, for each decoder query prediction $\mathbf{Q}^{pred}_{v} = \{\mathbf{Q}_{v}^{cls} , \mathbf{Q}_{v}^{box}\}$, we compute a pair-wise matching cost to ground-truths within the same image. The matching cost is a weighted sum of classification and localization terms:
\begin{equation}
\mathcal{C}^{i,j} = \lambda_{focal} \mathcal{C}_{focal}(\mathbf{q}^{pred_{i}}_{v}, \mathbf{g}^{j}) + \lambda_{l1} \mathcal{C}_{l1}(\mathbf{q}^{pred_{i}}_{v}, \mathbf{g}^{j}) + \lambda_{GIoU} \mathcal{C}_{GIoU}(\mathbf{q}^{pred_{i}}_{v}, \mathbf{g}^{j}).
\end{equation}
The minimum total cost among GT instances is selected as the metrics for evaluating the query quality. The reward $r_i$ for query $i$ is then defined as the inverse of this minimal cost:
\begin{equation}
\gamma_{i} = \min_{j}\mathcal{C}^{i,j}, \quad r_i = -\gamma_{i},
\end{equation}
where $C_{i,j}$ denotes the matching cost between query $\mathbf{q}^{pred_{i}}_{v}$ and ground-truth $\mathbf{g}^{j}$. A lower cost implies a better alignment, and thus a higher reward.
To make the learning signal more robust and harness the group dynamic, we normalize the reward across all queries within the same sample to compute a relative advantage:
\begin{equation}
\hat{A}_{i} = \frac{r_{i} - \mu_{r}}{\sigma_{r}}
\end{equation}
where $\mu_r$ and $\sigma_r$ are the mean and standard deviation of the rewards within the group. This group-normalized advantage provides stable, comparative gradients, encouraging all queries to improve relative to the dynamic group baseline.

\begin{figure}
    \centering
    \includegraphics[width=1.0\linewidth]{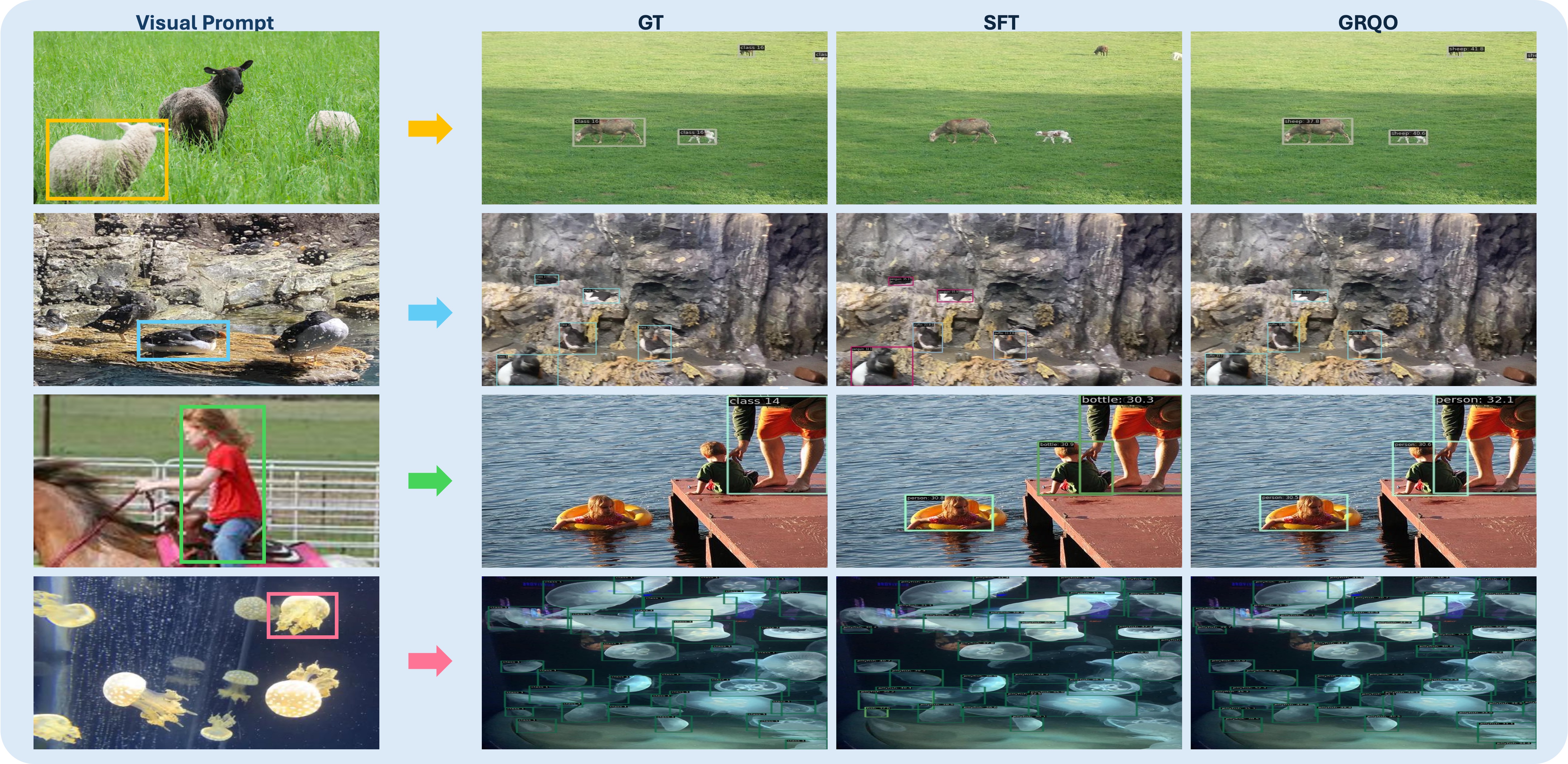}
    \caption{\textbf{Qualitative comparison of visual prompting detection between SFT and GRQO.} SFT results exhibit both false positives (row 2,3,4) and missed detections (row 1), reflecting limited query expressiveness and weak alignment with visual prompts. In contrast, GRQO produces more accurate and complete detections, better aligning with the prompted semantics. These results highlight GRQO’s ability to enhance query reasoning and robustness under high-variance visual inputs.}
    \label{fig:qualitative-cmp}
\end{figure}

\paragraph{KL-divergence regularization.}
To further stabilize training under high-variance visual prompts and prevent distributional drift, we introduce a KL-divergence-based regularization term on the objectness probability distribution. In our setting, the objectness distribution captures the model's confidence over image tokens being relevant to the prompted object. Due to the diverse appearance and structure of visual prompts, these objectness predictions can fluctuate across iterations, leading to training instability.
To mitigate this, we regularize the objectness probability distribution $\mathcal{O}_{\theta}$ of the current model with respect to a reference model distribution $\mathcal{O}_{ref}$. Given the selected top $N_q$ token indices, the two distribution is generated as Eq.:
\begin{equation}
\mathcal{O}_{\theta} = \mathrm{Max}^{(-1)}(\mathbf{I}' \cdot {\bar{\mathbf{v}}}^{'\top})[\mathrm{Idx}_{N_q}^{v}], \quad \mathcal{O}_{ref} = \mathrm{Max}^{(-1)}(\mathbf{I}'_{ref} \cdot {\bar{\mathbf{v}}}^{'\top}_{ref})[\mathrm{Idx}_{N_q}^{v}],
\end{equation}
where $\mathbf{I}'_{ref}$ and ${\bar{\mathbf{v}}}^{'\top}_{ref}$ denote the refined image features and prompt features from reference model. Then, the KL-divergence is computed as:
\begin{equation}
\mathcal{D}_{KL}[\mathcal{O}_{\theta}\parallel \mathcal{O}_{ref}] = \frac{\mathcal{O}_{ref}(q_{i}|\mathbf{I}, \bar{\mathbf{v}})}{\mathcal{O}_{\theta}(q_{i}|\mathbf{I}, \bar{\mathbf{v}})} - \mathrm{log} \frac{\mathcal{O}_{ref}(q_{i}|\mathbf{I}, \bar{\mathbf{v}})}{\mathcal{O}_{\theta}(q_{i}|\mathbf{I}, \bar{\mathbf{v}})} - 1,
\end{equation}
where $q_i$ denotes the $i$-th query token, $\mathcal{O}$ represents the query objectness distribution conditioned on target image features $\mathbf{I}$ and sampled visual prompts $\bar{\mathbf{v}}$.
This regularization encourages the current model to remain close to the reference distribution, which is a frozen copy of an earlier training state. By anchoring the learning dynamics to a stable prior, KL-regularization helps the model retain generalizable knowledge while progressively absorbing the diversity of visual prompts.

\paragraph{Overall training objective.}
Our proposed GRQO loss introduces group-relative reinforcement signals and regularization to improve query quality and training stability. Specifically, the GRQO loss is defined as:
\begin{equation}
\mathcal{L}_\text{GRQO} = -\frac{1}{N_q}\sum_{i=1}^{N_q}(\alpha \times \hat{A}_{i} - \beta \times \mathcal{D}_{KL}[\mathcal{O}_{\theta}\parallel \mathcal{O}_{ref}]),
\end{equation}
where $\alpha$ and $\beta$ are scalar weights that balance the reward signal and regularization strength. To mitigate the impact of unreliable signals, we set $\alpha$ to zero for the final set of queries exhibiting low objectness probability. GRQO incentivizes both query-level learning and stable objectness modeling.

To complement this group-level supervision, we include standard per-query detection losses.
In addition, we retain the region-level contrastive loss $\mathcal{L}_\text{contra}$ to align visual prompts with corresponding semantic concepts.
The final training objective of DINO-R1 is guided by a composite objective:
\begin{equation}
\mathcal{L}_\text{DINO-R1} = \mathcal{L}_\text{GRQO} + \mathcal{L}_\text{contra} + \mathcal{L}_\text{focal} + \mathcal{L}_{L1} + \mathcal{L}_\text{GIoU}.
\end{equation}
This multi-component loss encourages DINO-R1 to benefit from both group-level optimization signals and instance-level supervision, enabling robust and generalizable visual prompting detection.

\section{Experiments}
\label{sec:experiments}

\subsection{Experiment Setup}
\paragraph{Baseline and Base Model.} We compare GRQO with standard supervised fine-tuning (SFT). All experiments are conducted using the \texttt{MM-Grounding-DINO} \cite{mmgdino} implementation, which we adapt to support visual prompting. We use visual exemplars (images with user defined bounding boxes) as input prompts to guide the detection in the target image.

\paragraph{Datasets and Implementation Details.} We evaluate our approach under two settings:
\ding{182} \textbf{Zero-Shot. (Out-domain Evaluation).} We conduct open-vocabulary detection where we train on Objects365 (O365) \cite{o365} and test on COCO \cite{cocods}, LVIS-minival \cite{lvis}, ODinW13 \cite{glip}, and ODinW35 \cite{glip}. We train the model for 6 epochs as SFT baselines. For GRQO, we first train 1 epoch with SFT to obtain the reference model, then apply GRQO for 5 additional epochs.
\ding{183} \textbf{Fine-Tune. (In-domain Evaluation).} We fine-tune on the COCO training split for 12 epochs and evaluate on the COCO validation set. For GRQO, we apply both pretrained weights of SFT and GRQO for evaluation.

\subsection{Main Results}
The visual prompting object detection results are summarized in Table.\ref{tab:main}.

\begin{table}
    \caption{\small \bf Object detection results under Zero-Shot and Fine-Tuning settings across multiple datasets.}
    \label{tab:main}
    \centering
    \resizebox{1.0\linewidth}{!}{
    \begin{tabular}{l|c|c|c|cccc|c|c|c}
      \toprule
      \multirow{3}{*}{Model} & \multirow{3}{*}{Dataset} & \multirow{3}{*}{Epoch} & \multicolumn{7}{c|}{Zero-Shot} & Fine-Tune \\
       & & & COCO & \multicolumn{4}{c|}{LVIS-minival} & ODinW13 & ODinW35 & COCO \\
       & & & AP & AP & $AP_{r}$ & $AP_{c}$ & $AP_{f}$ & $AP_{avg}$ & $AP_{avg}$ & AP \\
       \midrule
       \multicolumn{11}{c}{Text Prompting} \\
       \midrule
       GDINO-T \cite{gdino} & O365,G,C & 30 & 48.4 & 28.8 & 18.8 & 24.2 & 34.7 & 51.4 & 22.7 & 58.1 \\
       MM-GDINO-T \cite{mmgdino} & O365,G & 30 & 50.4 & 35.7 & 28.1 & 30.2 & 42.0 & 45.3 & 20.2 & 58.2 \\
       MM-GDINO-L \cite{mmgdino} & O365v2,O,G & 42 & 53.0 & - & - & - & - & 75.3 & 35.2 & - \\
       \midrule
       \multicolumn{11}{c}{Visual Prompting} \\
       \midrule
       \rowcolor{ProcessBlue!35} \multicolumn{11}{l}{\small $\emph{SFT}$} \\
       VIS-GDINO-T & O365 & 6 & 19.9 & 15.3 & 11.8 & 15.6 & 17.1 & 4.6  & 2.6 & 32.5 \\
       VIS-GDINO-B & O365 & 6 & 24.8 & 17.1 & 12.5 & 16.4 & 18.9 & 11.2 & 6.7 & 38.1 \\
       VIS-GDINO-L & O365 & 6 & 26.1 & 18.4 & 15.9 & 17.8 & 20.3 & 15.3 & 8.2 & 39.2 \\
       \rowcolor{Lavender!70} \multicolumn{11}{l}{\small $\emph{GRQO}$} \\
       \textit{DINO-R1}-T & O365 & 1+5 & 24.0 & 17.4 & 14.1 & 17.9 & 20.8 & 5.0  & 3.0  & 37.2/39.6 \\
       \textit{DINO-R1}-B & O365 & 1+5 & 26.1 & 19.0 & 15.9 & 19.4 & 21.2 & 14.7 & 7.2  & 41.3/42.4 \\
       \textit{DINO-R1}-L & O365 & 1+5 & 28.1 & 20.1 & 17.8 & 19.8 & 23.4 & 24.1 & 12.6 & 43.5/44.1 \\
      \bottomrule
    \end{tabular}
    }
  \end{table}

\paragraph{Out-of-domain detection on COCO and LVIS.}
We evaluate models trained on Objects365 under zero-shot transfer settings. As shown in Table.\ref{tab:main}, DINO-R1 consistently achieves better generalization to COCO and LVIS datasets. On COCO, \textit{DINO-R1}-T improves mAP by +4.1 (19.9 $\rightarrow$ 24.0) compared to SFT. On the more challenging LVIS dataset, which contains long-tailed categories, \textit{DINO-R1}-B improves over SFT by +3.4 (12.5 $\rightarrow$ 15.9) on rare category, demonstrating its stronger generalization to diverse and rare categories. This validates the effectiveness of GRQO’s group-wise learning and regularization in handling open-vocabulary visual conditions. Fig.~\ref{fig:map-prompt-vis}(a) presents the training dynamics, where DINO-R1 exhibits more stable training compared to SFT.

\paragraph{Out-of-domain detection in the wild.}
We further evaluate our method on ODinW, which includes various real-world domains. On the 13-dataset and 35-dataset ODinW subsets, \textit{DINO-R1}-L outperforms SFT by +8.8 and +4.4 mAP, respectively.
These consistent gains across varied domains reflect not only improved generalization, but also enhanced visual reasoning capability. By optimizing queries with group-relative rewards and stable objectness supervision, \textit{DINO-R1} learns to better align high-level semantics across disparate scenes and object styles—an essential property for visual in-context reasoning in open-world scenarios. We provide qualitative comparison in Fig.\ref{fig:qualitative-cmp}.

\begin{table}[t]
  \centering
  \begin{minipage}[t]{0.38\linewidth}
  \caption{\textbf{Ablation of the query-level reward and KL-regularization modules.} Both individually improve performance over SFT, while combining them yields the highest gains in both settings on COCO.}
  \label{tab:abl-component}
    \resizebox{1.0\linewidth}{!}{
    \begin{tabular}{l|c|c}
    \toprule
    \multirow{3}{*}{Method} & \multicolumn{2}{c}{COCO} \\
     & Zero-Shot & Fine-Tune \\
     & AP & AP \\
     \midrule
     SFT         & 19.9 & 32.5 \\
     only reward & 22.8 & 36.1 \\
     only KL-Div & 21.0 & 34.2 \\
     GRQO        & 24.0 & 37.2 \\
    \bottomrule
  \end{tabular}
    }
    \end{minipage} \quad
  \begin{minipage}[t]{0.58\linewidth}
      \renewcommand{\arraystretch}{0.94}
  \caption{\textbf{Design analysis of the query reward.} Ablation of reward formulation using different combinations of focal, L1, and IoU-based costs. Group-relative rewards consistently outperform absolute variants, and layer-wise reward propagation ($\dagger$) further enhances performance.}
  \label{tab:abl-reward}
    \resizebox{1.0\linewidth}{!}{
   \begin{tabular}{l|ccc|c|c}
    \toprule
    \multirow{3}{*}{Method} & \multicolumn{3}{c|}{Reward} & \multicolumn{2}{c}{COCO} \\
     & \multirow{2}{*}{Focal} & \multirow{2}{*}{BboxL1} & \multirow{2}{*}{IoU} & Zero-Shot & Fine-Tune \\
     & & & & AP & AP \\
     \midrule
        SFT & - & - & - & 19.9 & 32.5 \\
        GRQO (relative) & & \checkmark & \checkmark & 21.3 & 34.1 \\
        GRQO (relative) & \checkmark & & \checkmark & 22.7 & 33.6 \\
        GRQO (relative) & \checkmark & \checkmark & & 21.8 & 34.0 \\
        GRQO (relative) & \checkmark & \checkmark & \checkmark & 23.5 & 36.8 \\
        GRQO (absolute) & \checkmark & \checkmark & \checkmark & 20.1 & 31.4 \\
        GRQO (relative) $\dagger$ & \checkmark & \checkmark & \checkmark & 24.0 & 37.2 \\
    \bottomrule
  \end{tabular}
    }
    \end{minipage}
    \vspace{-5pt}
\end{table}

\paragraph{In-domain detection on COCO.}
GRQO also provides consistent gains under the closed-set detection setting on COCO across multiple training strategies. When fine-tuning the SFT-pretrained model with GRQO, \textit{DINO-R1}-L achieves 43.5 mAP, outperforming continued SFT training (39.2 mAP) by +4.3. Notably, using the GRQO-pretrained model as the starting point leads to even larger improvements, with \textit{DINO-R1} surpassing the SFT baseline by +4.9 mAP. These results demonstrate that GRQO not only generalizes better but also improves training efficiency and effectiveness within the same domain.

\subsection{Ablation Study}

\paragraph{Effectiveness of each component.}
To evaluate the contributions of the two key components in GRQO—query-level relative reward and KL-divergence regularization—we conduct controlled ablations by enabling each module independently. Table~\ref{tab:abl-component} shows that both components individually improve performance over the SFT baseline for both out-domain and in-domain detection. Specifically, the reward module yields 2.9 and 3.6 mAP gains, while the KL-regularization contributes 1.1 and 1.7 mAP improvements. When both components are applied together, the full GRQO framework further outperforms the SFT baseline by 4.1 and 4.7, respectively. These results confirm that both modules are beneficial, and their combination further enhances the model’s capability to generalize under visual prompting settings.

\begin{wraptable}[15]{c}
{0.37\textwidth}
\centering
\vspace{-4.5mm}
\caption{
\small
\textbf{Impact of loss weights in GRQO.} We vary the scaling of the query-level reward and KL-regularization losses. The best performance is achieved with a reward weight of 10e3 and a KL weight of 0.04, highlighting the importance of balancing learning signal strength and regularization.}
\vspace{-1mm}
\label{tab:abl-lossweight}
\small
\resizebox{0.37\textwidth}{!}{
\begin{tabular}{cc|c|c}
    \toprule
     \multicolumn{2}{c|}{Loss Weight} & \multicolumn{2}{c}{COCO} \\
     \multirow{2}{*}{Reward} & \multirow{2}{*}{KL-Div} & Zero-Shot & Fine-Tune \\
      & & AP & AP \\
     \midrule
      1.0  & 0.4   & 20.2 & 33.4 \\
      1.0  & 0.04  & 21.6 & 35.2 \\
      10.0 & 0.04  & 22.4 & 35.1 \\
      10e3 & 0.04  & 24.0 & 37.2 \\
      10e4 & 0.04  & 23.1 & 36.8 \\
      10e3 & 0.004 & 21.5 & 35.3 \\
    \bottomrule
  \end{tabular}
}
\end{wraptable}

\paragraph{Query reward design.}
We ablate the design choices in the reward function used to optimize query quality. Since our goal is accurate detection guided by visual prompts, we consider both classification and localization cues for reward formulation. We test the classification reward (reverse focal cost), the localization reward (the reverse L1 and GIoU). Additionally, we compare the use of absolute reward values versus group-relative reward values.

As shown in Table.\ref{tab:abl-reward}, using all three components together with group-relative reward achieves the best performance of 23.5 and 36.8 mAP. Notably, the relative reward outperforms absolute reward by a margin of 3.4 and 5.4 mAP, highlighting that group-wise normalization improves reward stability and allows the model to focus on inter-query discriminability rather than absolute query quality, which is often sensitive to instance-level noise. Furthermore, we examine a layer-wise reward strategy, where intermediate decoder layers are also supervised by the reward function. As shown in the last row of Table.\ref{tab:abl-reward}, this design further boosts performance by 0.5 and 0.4 mAP, suggesting that earlier query refinement stages also benefit from reinforcement-style optimization.

\paragraph{Effect of Loss Scaling.}
We investigate the sensitivity of GRQO to the scaling of its two key loss components: the query reward term and the KL-divergence regularization. Specifically, we vary the weight of the reward loss across {1.0, 10.0, 10e2, 10e3, 10e4}, and the KL-regularization across {0.4, 0.04, 0.004}. As shown in Table~\ref{tab:abl-lossweight}, the best performance is achieved when the reward weight is set to 10e3 and the KL weight to 0.04. This indicates that moderately strong reward signals encourage more effective query discrimination, while excessively large weights lead to inferior optimization. Similarly, the KL-regularization coefficient of 0.04 strikes a good balance between stability and generalization, helping the model resist distributional drift during training on diverse visual prompts.

\paragraph{Number of prompts.}
The diversity and quantity of visual prompts play a crucial role in training robust visual-prompting detectors. We ablate the number of randomly sampled prompts per class during training, and further evaluate models with varying numbers of prompts at inference. As shown in Fig.\ref{fig:map-prompt-vis}(b)(c) and Table~\ref{tab:abl-numprompt}, training with only one random prompt per class significantly outperforms settings with more prompts. We hypothesize that this is due to the increased diversity and higher variance in the sampled prompt pool, allowing the model to generalize over a wider range of visual appearances. By seeing more varied exemplars across training iterations, the model learns a broader and more adaptable visual concept space. Conversely, during inference, the performance improves as the number of prompts per class increases, suggesting that ensemble-style prompting helps reinforce object identity and reduce ambiguity in open-set scenarios.

\begin{minipage}{1.0\textwidth}
\noindent
  \begin{minipage}[b]{0.76\linewidth}
    \centering
    \includegraphics[width=1\linewidth]{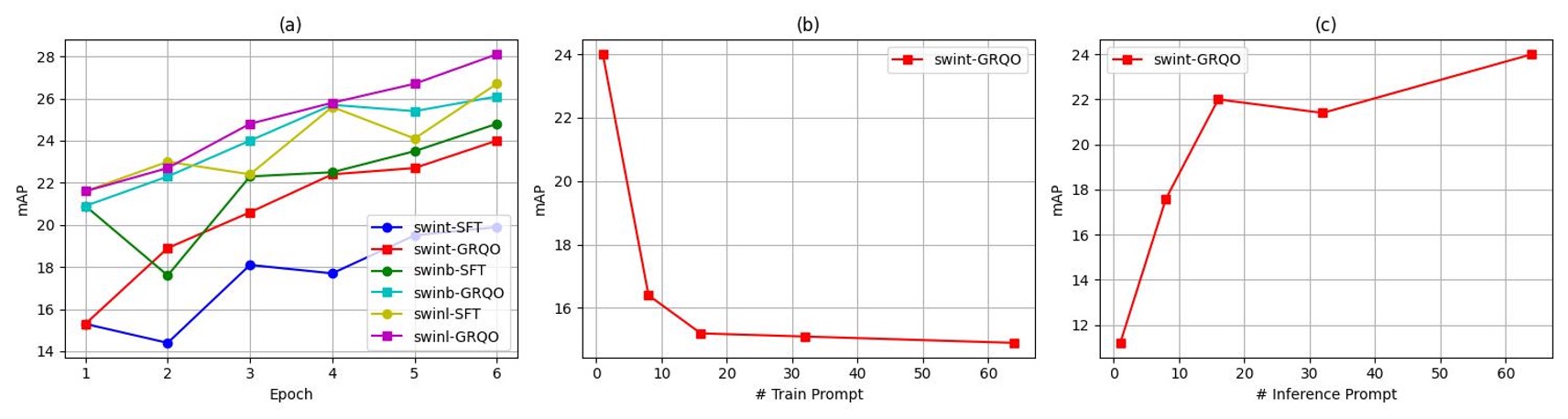}
    \captionof{figure}{{\small  (a) Training curves comparing SFT and GRQO.} GRQO consistently achieves more stable training with higher final performance. (b)(c) Effect of the number of prompts per class during training/inference.}
    \label{fig:map-prompt-vis}
  \end{minipage}
  \quad
  \begin{minipage}[b]{0.21\linewidth}
    \centering
     \resizebox{1.0\linewidth}{!}{
     \begin{tabular}{c|c|c}
    \hline
     \multirow{2}{*}{\#} & \multicolumn{2}{c}{COCO Zero-Shot} \\
       & Train & Inference \\
     \toprule
     1  & \textbf{24.0} & 11.2 \\
     8  & 16.4 & 17.6 \\
     16 & 15.2 & 22.0 \\
     32 & 15.1 & 21.4 \\
     64 & 14.9 & \textbf{24.0} \\
    \bottomrule
  \end{tabular}
    }
      \captionof{table}{\small {Effect of the number of prompts.} We vary the number of sampled prompts per class during training and inference.} \label{tab:abl-numprompt}
    \end{minipage}
  \end{minipage}

\section{Conclusion}
\label{sec:conclusion}
\vspace{-2mm}
We introduce \textit{DINO-R1}, a novel training paradigm that advances the reasoning capabilities of vision foundation models under the visual prompting setting. Built on Grounding DINO, DINO-R1 rethinks how object queries are trained by shifting from sparse, instance-level supervision to dense, group-aware optimization. At the heart of our method is Group Relative Query Optimization (GRQO), which evaluates and refines queries through relative rewards within query groups—mirroring the collaborative inference process inherent in transformer-based detectors. To further stabilize training and prevent forgetting, we propose a KL-divergence regularization on the objectness distribution, anchoring the model around stable representations while progressively learning from diverse prompts. Together, these components offer a principled and scalable approach for training detectors that generalize robustly across domains, exhibit stronger reasoning over visual prompts, and remain stable under the high variance characteristic of open-world conditions. Our extensive experiments on COCO, LVIS, and ODinW validate the effectiveness of DINO-R1, showing consistent improvements over supervised fine-tuning across both zero-shot and fine-tuned evaluations.

We believe DINO-R1 opens a promising direction for \textbf{reinforcement-inspired training} in dense visual tasks and provides a foundation for future research in visual in-context learning, multi-modal alignment, and prompt-driven visual reasoning.
\vspace{-2mm}

\section{Future Work \& Limitations}
\label{sec:limitation}
\vspace{-2mm}
Our work primarily focuses on the optimization strategy rather than architectural enhancements. The visual prompt encoder used in \textit{DINO-R1} adopts a relatively simple design to isolate and highlight the contributions of our GRQO framework. We believe there is substantial room to explore more expressive and structured visual prompt encoding methods. In future work, we plan to integrate advanced visual prompt architectures, extend \textit{DINO-R1} to more challenging and diverse datasets, and explore its application in other open-world settings such as referring expression comprehension, retrieval-augmented detection, and multi-shot visual reasoning. We see \textit{DINO-R1} as a foundational step toward scalable, prompt-driven visual understanding—and aim to build on this foundation by further closing the gap between model flexibility and reasoning robustness.

%
%
\medskip


{
    \small
    \bibliographystyle{unsrt}
    \bibliography{ref}

\begin{thebibliography}{10}

\bibitem{grpo}
Zhihong Shao, Peiyi Wang, Qihao Zhu, Runxin Xu, Junxiao Song, Xiao Bi, Haowei Zhang, Mingchuan Zhang, YK~Li, Y~Wu, et~al.
\newblock Deepseekmath: Pushing the limits of mathematical reasoning in open language models.
\newblock {\em arXiv preprint arXiv:2402.03300}, 2024.

\bibitem{llama2}
Hugo Touvron, Louis Martin, Kevin Stone, Peter Albert, Amjad Almahairi, Yasmine Babaei, Nikolay Bashlykov, Soumya Batra, Prajjwal Bhargava, Shruti Bhosale, et~al.
\newblock Llama 2: Open foundation and fine-tuned chat models.
\newblock {\em arXiv preprint arXiv:2307.09288}, 2023.

\bibitem{llama3}
Aaron Grattafiori, Abhimanyu Dubey, Abhinav Jauhri, Abhinav Pandey, Abhishek Kadian, Ahmad Al-Dahle, Aiesha Letman, Akhil Mathur, Alan Schelten, Alex Vaughan, et~al.
\newblock The llama 3 herd of models.
\newblock {\em arXiv preprint arXiv:2407.21783}, 2024.

\bibitem{qwen}
Jinze Bai, Shuai Bai, Yunfei Chu, Zeyu Cui, Kai Dang, Xiaodong Deng, Yang Fan, Wenbin Ge, Yu~Han, Fei Huang, et~al.
\newblock Qwen technical report.
\newblock {\em arXiv preprint arXiv:2309.16609}, 2023.

\bibitem{qwen2}
An~Yang, Baosong Yang, Beichen Zhang, Binyuan Hui, Bo~Zheng, Bowen Yu, Chengyuan Li, Dayiheng Liu, Fei Huang, Haoran Wei, et~al.
\newblock Qwen2. 5 technical report.
\newblock {\em arXiv preprint arXiv:2412.15115}, 2024.

\bibitem{bi2024deepseek}
Xiao Bi, Deli Chen, Guanting Chen, Shanhuang Chen, Damai Dai, Chengqi Deng, Honghui Ding, Kai Dong, Qiushi Du, Zhe Fu, et~al.
\newblock Deepseek llm: Scaling open-source language models with longtermism.
\newblock {\em arXiv preprint arXiv:2401.02954}, 2024.

\bibitem{lu2024deepseek}
Haoyu Lu, Wen Liu, Bo~Zhang, Bingxuan Wang, Kai Dong, Bo~Liu, Jingxiang Sun, Tongzheng Ren, Zhuoshu Li, Hao Yang, et~al.
\newblock Deepseek-vl: towards real-world vision-language understanding.
\newblock {\em arXiv preprint arXiv:2403.05525}, 2024.

\bibitem{guo2025deepseek}
Daya Guo, Dejian Yang, Haowei Zhang, Junxiao Song, Ruoyu Zhang, Runxin Xu, Qihao Zhu, Shirong Ma, Peiyi Wang, Xiao Bi, et~al.
\newblock Deepseek-r1: Incentivizing reasoning capability in llms via reinforcement learning.
\newblock {\em arXiv preprint arXiv:2501.12948}, 2025.

\bibitem{liu2024deepseek}
Aixin Liu, Bei Feng, Bing Xue, Bingxuan Wang, Bochao Wu, Chengda Lu, Chenggang Zhao, Chengqi Deng, Chenyu Zhang, Chong Ruan, et~al.
\newblock Deepseek-v3 technical report.
\newblock {\em arXiv preprint arXiv:2412.19437}, 2024.

\bibitem{llava}
Haotian Liu, Chunyuan Li, Qingyang Wu, and Yong~Jae Lee.
\newblock Visual instruction tuning.
\newblock {\em Advances in neural information processing systems}, 36:34892--34916, 2023.

\bibitem{sam}
Alexander Kirillov, Eric Mintun, Nikhila Ravi, Hanzi Mao, Chloe Rolland, Laura Gustafson, Tete Xiao, Spencer Whitehead, Alexander~C Berg, Wan-Yen Lo, et~al.
\newblock Segment anything.
\newblock In {\em Proceedings of the IEEE/CVF international conference on computer vision}, pages 4015--4026, 2023.

\bibitem{sam2}
Nikhila Ravi, Valentin Gabeur, Yuan-Ting Hu, Ronghang Hu, Chaitanya Ryali, Tengyu Ma, Haitham Khedr, Roman R{\"a}dle, Chloe Rolland, Laura Gustafson, et~al.
\newblock Sam 2: Segment anything in images and videos.
\newblock {\em arXiv preprint arXiv:2408.00714}, 2024.

\bibitem{dino}
Hao Zhang, Feng Li, Shilong Liu, Lei Zhang, Hang Su, Jun Zhu, Lionel~M Ni, and Heung-Yeung Shum.
\newblock Dino: Detr with improved denoising anchor boxes for end-to-end object detection.
\newblock {\em arXiv preprint arXiv:2203.03605}, 2022.

\bibitem{gdino}
Shilong Liu, Zhaoyang Zeng, Tianhe Ren, Feng Li, Hao Zhang, Jie Yang, Qing Jiang, Chunyuan Li, Jianwei Yang, Hang Su, et~al.
\newblock Grounding dino: Marrying dino with grounded pre-training for open-set object detection.
\newblock In {\em European Conference on Computer Vision}, pages 38--55. Springer, 2024.

\bibitem{dinox}
Tianhe Ren, Yihao Chen, Qing Jiang, Zhaoyang Zeng, Yuda Xiong, Wenlong Liu, Zhengyu Ma, Junyi Shen, Yuan Gao, Xiaoke Jiang, et~al.
\newblock Dino-x: A unified vision model for open-world object detection and understanding.
\newblock {\em arXiv preprint arXiv:2411.14347}, 2024.

\bibitem{dinov}
Feng Li, Qing Jiang, Hao Zhang, Tianhe Ren, Shilong Liu, Xueyan Zou, Huaizhe Xu, Hongyang Li, Jianwei Yang, Chunyuan Li, et~al.
\newblock Visual in-context prompting.
\newblock In {\em Proceedings of the IEEE/CVF Conference on Computer Vision and Pattern Recognition}, pages 12861--12871, 2024.

\bibitem{vit}
Alexey Dosovitskiy, Lucas Beyer, Alexander Kolesnikov, Dirk Weissenborn, Xiaohua Zhai, Thomas Unterthiner, Mostafa Dehghani, Matthias Minderer, Georg Heigold, Sylvain Gelly, et~al.
\newblock An image is worth 16x16 words: Transformers for image recognition at scale.
\newblock {\em arXiv preprint arXiv:2010.11929}, 2020.

\bibitem{imagenet}
Jia Deng, Wei Dong, Richard Socher, Li-Jia Li, Kai Li, and Li~Fei-Fei.
\newblock Imagenet: A large-scale hierarchical image database.
\newblock In {\em 2009 IEEE conference on computer vision and pattern recognition}, pages 248--255. Ieee, 2009.

\bibitem{cocods}
Tsung-Yi Lin, Michael Maire, Serge Belongie, James Hays, Pietro Perona, Deva Ramanan, Piotr Doll{\'a}r, and C~Lawrence Zitnick.
\newblock Microsoft coco: Common objects in context.
\newblock In {\em Computer vision--ECCV 2014: 13th European conference, zurich, Switzerland, September 6-12, 2014, proceedings, part v 13}, pages 740--755. Springer, 2014.

\bibitem{o365}
Shuai Shao, Zeming Li, Tianyuan Zhang, Chao Peng, Gang Yu, Xiangyu Zhang, Jing Li, and Jian Sun.
\newblock Objects365: A large-scale, high-quality dataset for object detection.
\newblock In {\em Proceedings of the IEEE/CVF international conference on computer vision}, pages 8430--8439, 2019.

\bibitem{dinov2}
Maxime Oquab, Timoth{\'e}e Darcet, Th{\'e}o Moutakanni, Huy Vo, Marc Szafraniec, Vasil Khalidov, Pierre Fernandez, Daniel Haziza, Francisco Massa, Alaaeldin El-Nouby, et~al.
\newblock Dinov2: Learning robust visual features without supervision.
\newblock {\em arXiv preprint arXiv:2304.07193}, 2023.

\bibitem{SimCLR}
Ting Chen, Simon Kornblith, Mohammad Norouzi, and Geoffrey Hinton.
\newblock A simple framework for contrastive learning of visual representations.
\newblock In {\em International conference on machine learning}, pages 1597--1607. PmLR, 2020.

\bibitem{G-SimCLR}
Souradip Chakraborty, Aritra~Roy Gosthipaty, and Sayak Paul.
\newblock G-simclr: Self-supervised contrastive learning with guided projection via pseudo labelling.
\newblock In {\em 2020 international conference on data mining workshops (ICDMW)}, pages 912--916. IEEE, 2020.

\bibitem{trex}
Qing Jiang, Feng Li, Tianhe Ren, Shilong Liu, Zhaoyang Zeng, Kent Yu, and Lei Zhang.
\newblock T-rex: Counting by visual prompting.
\newblock {\em arXiv preprint arXiv:2311.13596}, 2023.

\bibitem{trex2}
Qing Jiang, Feng Li, Zhaoyang Zeng, Tianhe Ren, Shilong Liu, and Lei Zhang.
\newblock T-rex2: Towards generic object detection via text-visual prompt synergy.
\newblock In {\em European Conference on Computer Vision}, pages 38--57. Springer, 2024.

\bibitem{cp-detr}
Qibo Chen, Weizhong Jin, Jianyue Ge, Mengdi Liu, Yuchao Yan, Jian Jiang, Li~Yu, Xuanjiang Guo, Shuchang Li, and Jianzhong Chen.
\newblock Cp-detr: Concept prompt guide detr toward stronger universal object detection.
\newblock In {\em Proceedings of the AAAI Conference on Artificial Intelligence}, volume~39, pages 2141--2149, 2025.

\bibitem{app-visual-autovp}
Hsi-Ai Tsao, Lei Hsiung, Pin-Yu Chen, Sijia Liu, and Tsung-Yi Ho.
\newblock Autovp: An automated visual prompting framework and benchmark.
\newblock {\em arXiv preprint arXiv:2310.08381}, 2023.

\bibitem{app-visual-moka}
Fangchen Liu, Kuan Fang, Pieter Abbeel, and Sergey Levine.
\newblock Moka: Open-vocabulary robotic manipulation through mark-based visual prompting.
\newblock In {\em First Workshop on Vision-Language Models for Navigation and Manipulation at ICRA 2024}, 2024.

\bibitem{app-visual-multimodal}
Yuang Ai, Huaibo Huang, Xiaoqiang Zhou, Jiexiang Wang, and Ran He.
\newblock Multimodal prompt perceiver: Empower adaptiveness generalizability and fidelity for all-in-one image restoration.
\newblock In {\em Proceedings of the IEEE/CVF Conference on Computer Vision and Pattern Recognition}, pages 25432--25444, 2024.

\bibitem{app-visual-pivot}
Soroush Nasiriany, Fei Xia, Wenhao Yu, Ted Xiao, Jacky Liang, Ishita Dasgupta, Annie Xie, Danny Driess, Ayzaan Wahid, Zhuo Xu, et~al.
\newblock Pivot: Iterative visual prompting elicits actionable knowledge for vlms.
\newblock {\em arXiv preprint arXiv:2402.07872}, 2024.

\bibitem{app-visual-prompt-inds}
Gunwoo Yong, Kahyun Jeon, Daeyoung Gil, and Ghang Lee.
\newblock Prompt engineering for zero-shot and few-shot defect detection and classification using a visual-language pretrained model.
\newblock {\em Computer-Aided Civil and Infrastructure Engineering}, 38(11):1536--1554, 2023.

\bibitem{app-visual-promptcharm}
Zhijie Wang, Yuheng Huang, Da~Song, Lei Ma, and Tianyi Zhang.
\newblock Promptcharm: Text-to-image generation through multi-modal prompting and refinement.
\newblock In {\em Proceedings of the 2024 CHI Conference on Human Factors in Computing Systems}, pages 1--21, 2024.

\bibitem{app-visual-prompting-inds}
jinheng zhou, Wu~Liu, Guang Yang, He~Zhao, and feiniu yuan.
\newblock Prompting industrial anomaly segment with large vision-language models.
\newblock In {\em Proceedings of the 6th ACM International Conference on Multimedia in Asia}, pages 1--1, 2024.

\bibitem{app-visual-survey}
Junda Wu, Zhehao Zhang, Yu~Xia, Xintong Li, Zhaoyang Xia, Aaron Chang, Tong Yu, Sungchul Kim, Ryan~A Rossi, Ruiyi Zhang, et~al.
\newblock Visual prompting in multimodal large language models: A survey.
\newblock {\em arXiv preprint arXiv:2409.15310}, 2024.

\bibitem{app-visual-understanding}
Aochuan Chen, Yuguang Yao, Pin-Yu Chen, Yihua Zhang, and Sijia Liu.
\newblock Understanding and improving visual prompting: A label-mapping perspective.
\newblock In {\em Proceedings of the IEEE/CVF Conference on Computer Vision and Pattern Recognition}, pages 19133--19143, 2023.

\bibitem{app-visual-vp3d}
Yang Chen, Yingwei Pan, Haibo Yang, Ting Yao, and Tao Mei.
\newblock Vp3d: Unleashing 2d visual prompt for text-to-3d generation.
\newblock In {\em Proceedings of the IEEE/CVF Conference on Computer Vision and Pattern Recognition}, pages 4896--4905, 2024.

\bibitem{gdino1.5}
Tianhe Ren, Qing Jiang, Shilong Liu, Zhaoyang Zeng, Wenlong Liu, Han Gao, Hongjie Huang, Zhengyu Ma, Xiaoke Jiang, Yihao Chen, et~al.
\newblock Grounding dino 1.5: Advance the" edge" of open-set object detection.
\newblock {\em arXiv preprint arXiv:2405.10300}, 2024.

\bibitem{mmgdino}
Xiangyu Zhao, Yicheng Chen, Shilin Xu, Xiangtai Li, Xinjiang Wang, Yining Li, and Haian Huang.
\newblock An open and comprehensive pipeline for unified object grounding and detection.
\newblock {\em arXiv preprint arXiv:2401.02361}, 2024.

\bibitem{ovd-prompt}
Yu~Du, Fangyun Wei, Zihe Zhang, Miaojing Shi, Yue Gao, and Guoqi Li.
\newblock Learning to prompt for open-vocabulary object detection with vision-language model.
\newblock In {\em Proceedings of the IEEE/CVF conference on computer vision and pattern recognition}, pages 14084--14093, 2022.

\bibitem{ppo}
John Schulman, Filip Wolski, Prafulla Dhariwal, Alec Radford, and Oleg Klimov.
\newblock Proximal policy optimization algorithms.
\newblock {\em arXiv preprint arXiv:1707.06347}, 2017.

\bibitem{rlhf}
Long Ouyang, Jeffrey Wu, Xu~Jiang, Diogo Almeida, Carroll Wainwright, Pamela Mishkin, Chong Zhang, Sandhini Agarwal, Katarina Slama, Alex Ray, et~al.
\newblock Training language models to follow instructions with human feedback.
\newblock {\em Advances in neural information processing systems}, 35:27730--27744, 2022.

\bibitem{dpo}
Rafael Rafailov, Archit Sharma, Eric Mitchell, Christopher~D Manning, Stefano Ermon, and Chelsea Finn.
\newblock Direct preference optimization: Your language model is secretly a reward model.
\newblock {\em Advances in Neural Information Processing Systems}, 36:53728--53741, 2023.

\bibitem{clip}
Alec Radford, Jong~Wook Kim, Chris Hallacy, Aditya Ramesh, Gabriel Goh, Sandhini Agarwal, Girish Sastry, Amanda Askell, Pamela Mishkin, Jack Clark, et~al.
\newblock Learning transferable visual models from natural language supervision.
\newblock In {\em International conference on machine learning}, pages 8748--8763. PmLR, 2021.

\bibitem{florence}
Lu~Yuan, Dongdong Chen, Yi-Ling Chen, Noel Codella, Xiyang Dai, Jianfeng Gao, Houdong Hu, Xuedong Huang, Boxin Li, Chunyuan Li, et~al.
\newblock Florence: A new foundation model for computer vision.
\newblock {\em arXiv preprint arXiv:2111.11432}, 2021.

\bibitem{glip}
Liunian~Harold Li, Pengchuan Zhang, Haotian Zhang, Jianwei Yang, Chunyuan Li, Yiwu Zhong, Lijuan Wang, Lu~Yuan, Lei Zhang, Jenq-Neng Hwang, et~al.
\newblock Grounded language-image pre-training.
\newblock In {\em Proceedings of the IEEE/CVF conference on computer vision and pattern recognition}, pages 10965--10975, 2022.

\bibitem{internvl}
Zhe Chen, Jiannan Wu, Wenhai Wang, Weijie Su, Guo Chen, Sen Xing, Muyan Zhong, Qinglong Zhang, Xizhou Zhu, Lewei Lu, et~al.
\newblock Internvl: Scaling up vision foundation models and aligning for generic visual-linguistic tasks.
\newblock In {\em Proceedings of the IEEE/CVF conference on computer vision and pattern recognition}, pages 24185--24198, 2024.

\bibitem{swin}
Ze~Liu, Yutong Lin, Yue Cao, Han Hu, Yixuan Wei, Zheng Zhang, Stephen Lin, and Baining Guo.
\newblock Swin transformer: Hierarchical vision transformer using shifted windows.
\newblock In {\em Proceedings of the IEEE/CVF international conference on computer vision}, pages 10012--10022, 2021.

\bibitem{vlmr1}
Haozhan Shen, Peng Liu, Jingcheng Li, Chunxin Fang, Yibo Ma, Jiajia Liao, Qiaoli Shen, Zilun Zhang, Kangjia Zhao, Qianqian Zhang, et~al.
\newblock Vlm-r1: A stable and generalizable r1-style large vision-language model.
\newblock {\em arXiv preprint arXiv:2504.07615}, 2025.

\bibitem{yoloworld}
Tianheng Cheng, Lin Song, Yixiao Ge, Wenyu Liu, Xinggang Wang, and Ying Shan.
\newblock Yolo-world: Real-time open-vocabulary object detection.
\newblock In {\em Proceedings of the IEEE/CVF Conference on Computer Vision and Pattern Recognition}, pages 16901--16911, 2024.

\bibitem{detr}
Nicolas Carion, Francisco Massa, Gabriel Synnaeve, Nicolas Usunier, Alexander Kirillov, and Sergey Zagoruyko.
\newblock End-to-end object detection with transformers.
\newblock In {\em European conference on computer vision}, pages 213--229. Springer, 2020.

\bibitem{dndetr}
Feng Li, Hao Zhang, Shilong Liu, Jian Guo, Lionel~M Ni, and Lei Zhang.
\newblock Dn-detr: Accelerate detr training by introducing query denoising.
\newblock In {\em Proceedings of the IEEE/CVF conference on computer vision and pattern recognition}, pages 13619--13627, 2022.

\bibitem{dabdetr}
Shilong Liu, Feng Li, Hao Zhang, Xiao Yang, Xianbiao Qi, Hang Su, Jun Zhu, and Lei Zhang.
\newblock Dab-detr: Dynamic anchor boxes are better queries for detr.
\newblock {\em arXiv preprint arXiv:2201.12329}, 2022.

\bibitem{ovdetr-cm}
Yuhang Zang, Wei Li, Kaiyang Zhou, Chen Huang, and Chen~Change Loy.
\newblock Open-vocabulary detr with conditional matching.
\newblock In {\em European conference on computer vision}, pages 106--122. Springer, 2022.

\bibitem{ovrcnn}
Alireza Zareian, Kevin~Dela Rosa, Derek~Hao Hu, and Shih-Fu Chang.
\newblock Open-vocabulary object detection using captions.
\newblock In {\em Proceedings of the IEEE/CVF conference on computer vision and pattern recognition}, pages 14393--14402, 2021.

\bibitem{ovd-distill}
Xiuye Gu, Tsung-Yi Lin, Weicheng Kuo, and Yin Cui.
\newblock Open-vocabulary object detection via vision and language knowledge distillation.
\newblock {\em arXiv preprint arXiv:2104.13921}, 2021.

\bibitem{ovseg-diff}
Jiarui Xu, Sifei Liu, Arash Vahdat, Wonmin Byeon, Xiaolong Wang, and Shalini De~Mello.
\newblock Open-vocabulary panoptic segmentation with text-to-image diffusion models.
\newblock In {\em Proceedings of the IEEE/CVF Conference on Computer Vision and Pattern Recognition}, pages 2955--2966, 2023.

\bibitem{bert}
Jacob Devlin, Ming-Wei Chang, Kenton Lee, and Kristina Toutanova.
\newblock Bert: Pre-training of deep bidirectional transformers for language understanding.
\newblock In {\em Proceedings of the 2019 conference of the North American chapter of the association for computational linguistics: human language technologies, volume 1 (long and short papers)}, pages 4171--4186, 2019.

\bibitem{lvis}
Agrim Gupta, Piotr Dollar, and Ross Girshick.
\newblock Lvis: A dataset for large vocabulary instance segmentation.
\newblock In {\em Proceedings of the IEEE/CVF conference on computer vision and pattern recognition}, pages 5356--5364, 2019.

\end{thebibliography}
}

%
%
%
%





\end{document}